\pdfoutput=1
\documentclass[11pt]{article}
\usepackage[acronym]{glossaries}
\usepackage{graphicx}
\usepackage{paralist}

\usepackage[]{ACL2023}

\usepackage{times}
\usepackage{latexsym}

\usepackage[T1]{fontenc}
\usepackage[utf8]{inputenc}

\usepackage{microtype}

\usepackage{inconsolata}

\title{Data Bias According to Bipol: Men are Naturally Right and It is the Role of Women to Follow Their Lead}

\author{Irene Pagliai\textsuperscript{*1}, Goya van Boven\textsuperscript{*2}, Tosin Adewumi\textsuperscript{*$\dag$}, Lama Alkhaled\textsuperscript{$\dag$}, Namrata Gurung\textsuperscript{3},\\ \textbf{Isabella Södergren}\textsuperscript{4} and \textbf{Elisa Barney}\textsuperscript{$\dag$}  \\
\textsuperscript{1}University of Göttingen, Germany, \textsuperscript{2}Utrecht University, the Netherlands, \\
\textsuperscript{*$\dag$}Machine Learning Group, LTU, Sweden, \textsuperscript{3}QualityMinds GmbH, Germany,
\\ \textsuperscript{4}Digital Services and Systems, LTU, Sweden \\\textsuperscript{1}irene.pagliai@uni-goettingen.de, \textsuperscript{2}j.g.vanboven@students.uu.nl, \textsuperscript{$\dag$}firstname.lastname@ltu.se,\\ 
\textsuperscript{3}namrata.gurung@qualityminds.de \textsuperscript{4}isasde-5@student.ltu.se | \textsuperscript{*}\footnotesize{Joint first authors}\\
}

\makeglossaries
\newacronym{ood}{OOD}{out-of-distribution}
\newacronym{nlp}{NLP}{Natural Language Processing}
\newacronym{ner}{NER}{Named Entity Recognition}
\newacronym{sa}{SA}{Sentiment Analysis}
\newacronym{bow}{BoW}{bag-of-words}
\newacronym{cbow}{CBoW}{continuous Bag-of-Words}
\newacronym{sltc}{SLTC}{Swedish Language Technology Conference}
\newacronym{ann}{ANN}{artificial neural network}
\newacronym{nn}{NN}{neural network}
\newacronym{lstm}{LSTM}{Long Short Term Memory Network}
\newacronym{sota}{SotA}{state-of-the-art}
\newacronym{nlg}{NLG}{Natural Language Generation}
\newacronym{mwe}{MWE}{Multi-Word Expression}
\newacronym{cnn}{CNN}{Convolutional Neural Network}
\newacronym{sw}{SW}{Simple Wiki}
\newacronym{mt}{MT}{Machine Translation}
\newacronym{gdc}{GDC}{Gothenburg Dialog Corpus}
\newacronym{t5}{T5}{Text-to-Text-Transfer Transformer}
\newacronym{roberta}{RoBERTa}{Robustly optimized BERT approach}
\newacronym{bert}{BERT}{Bidirectional Encoder Representations from Transformers}
\newacronym{mcc}{MCC}{Matthews Correlation Coefficient}
\newacronym{ai}{AI}{artificial intelligence}
\newacronym{xai}{XAI}{explainable artificial intelligence}
\newacronym{lime}{LIME}{Local Interpretable Model-agnostic Explanations}
\newacronym{bilstm}{Bi-LSTM}{Bi- Directional Long Short Term Memory Network}
\newacronym{rnn}{RNN}{Recurrent Neural Network}
\newacronym{ml}{ML}{machine learning}
\newacronym{hasoc}{HASOC}{hate speech and offensive content}
\newacronym{olid}{OLID}{offensive language identification dataset}
\newacronym{ig}{IG}{Integrated Gradient}
\newacronym{shap}{SHAP}{SHapley Additive exPlanations}
\newacronym{hs}{HS}{hate speech}
\newacronym{trac}{TRAC}{Trolling, Aggression and Cyberbullying}
\newacronym{hso}{HSO}{hate speech and offensive}
\newacronym{ooc}{OOC}{out-of-class}
\newacronym{os}{OS}{operating system}
\newacronym{lr}{LR}{learning rate}
\newacronym{multiwoz}{MultiWOZ}{Multi-Domain Wizard-of-Oz}
\newacronym{dialogpt}{DialoGPT}{Dialogue Generative Pre-trained Transformer}
\newacronym{bold}{BOLD}{Bias in Open-Ended Language Generation Dataset}
\newacronym{cuad}{CUAD}{Contract Understanding Atticus Dataset}
\newacronym{ledgar}{LEDGAR}{Labeled Electronic Data Gathering, Analysis, and Retrieval system}
\newacronym{genbit}{GenBiT}{
Gender bias in text toolkit}
\newacronym{imdb}{IMDB}{
Internet Movie Database}
\newacronym{mdgender}{MDGender}{
Multi-Dimensional Gender}
\newacronym{sbic}{SBICv2}{Social Bias Inference Corpus v2}
\newacronym{tp}{tp}{true positives}
\newacronym{tn}{tn}{true negatives}
\newacronym{fp}{fp}{false positives}
\newacronym{fn}{fn}{false negatives}
\newacronym{bat}{BAT}{Bias-Aware Thresholding}
\newacronym{tf}{TF}{term frequency}
\newacronym{mab}{MAB}{multi-axes bias dataset}
\newacronym{squad}{SQuAD}{Stanford Question Answering Dataset}
\newacronym{copa}{COPA}{Choice Of Plausible Alternatives}
\newacronym{oodo}{OOD}{out-of-domain}
\newacronym{pos}{PoS}{part-of-speech}
\newacronym{pii}{PII}{personal identifiable information}
\newacronym{llm}{LLM}{large language model}
\newacronym{llms}{LLMs}{large language models}
\newacronym{mt5}{mT5}{multilingual Text-to-Text Transfer Transformer}
\newacronym{cb}{CB}{CommitmentBank}
\newacronym{mturk}{MTurk}{Amazon Mechanical Turk}
\newacronym{iaa}{IAA}{inter-annotator agreement}
\newacronym{nmt}{NMT}{neural machine translation}
\newacronym{cola}{CoLA}{Corpus of Linguistic Acceptability}
\newacronym{qnli}{QNLI}{Question-Answering Natural Language Inference}
\newacronym{mnli}{MNLI}{Multi-Genre  Natural Language Inference}
\newacronym{mrpc}{MRPC}{Microsoft Research Paraphrase Corpus}
\newacronym{stsb}{STSB}{Semantic Textual Similarity Benchmark}
\newacronym{record}{ReCoRD}{Reading Comprehension with Commonsense Reasoning Dataset}
\newacronym{gnad10}{GNAD10}{German News Articles Datasets 10k}
\newacronym{cus}{CUS}{credibility unanimous score}

\begin{document}
\maketitle
\begin{abstract}

We introduce new large labeled datasets on bias in 3 languages and show in experiments that bias exists in all 10 datasets of 5 languages evaluated, including benchmark datasets on the English GLUE/SuperGLUE leaderboards.
The 3 new languages give a total of almost 6 million labeled samples and we benchmark on these datasets using \acrshort{sota} multilingual pretrained models: \acrshort{mt5} and m\acrshort{bert}.
The challenge of social bias, based on prejudice, is ubiquitous, as recent events with \acrshort{ai} and \acrfull{llms} have shown.
Motivated by this challenge, we set out to estimate bias in multiple datasets.
We compare some recent bias metrics and use bipol, which has explainability in the metric.
We also confirm the unverified assumption that bias exists in toxic comments by randomly sampling 200 samples from a toxic dataset population using the confidence level of 95\% and error margin of 7\%.
Thirty gold samples were randomly distributed in the 200 samples to secure the quality of the annotation.
Our findings confirm that many of the datasets have male bias (prejudice against women), besides other types of bias.
We publicly release our new datasets, lexica, models, and codes.

\end{abstract}

%\textbf{Caution: This paper contains examples of what some may consider as stereotypes or offensive text.}

\section{Introduction}
The problem of social bias in data is a pressing one.
Recent news about social bias of \acrfull{ai} systems, such as Alexa\footnote{bbc.com/news/technology-66508514} and ChatGPT,\footnote{bloomberg.com/news/newsletters/2022-12-08/chatgpt-open-ai-s-chatbot-is-spitting-out-biased-sexist-results} shows that the age-old problem persists with data, which is used to train \acrfull{ml} models.
Social bias is the inclination or prejudice for, or against, a person, group or idea, especially in a way that is considered to be unfair, which may be based on race, religion or other factors \citep{bellamy2018ai,antoniak2021bad,mehrabi2021survey,ALKHALED2023100030}.
It can also involve stereotypes that generalize behavior to groups \citep{sep-implicit-bias}.
It can unfairly skew the output of \acrshort{ml} models \citep{klare2012face,10.1145/3375627.3375820}.
Languages with fewer resources than English are also affected \citep{rescigno2020case,chavez-mulsa-spanakis-2020-evaluating,kurpicz2020cultural}.
For example, in Italian, the female gender is under-represented due to the phenomena such as the ``inclusive masculine" (when the masculine is over-extended to denote groups of both male and female referents)
%and professional nouns are used in the masculine form, even when the person practicing the profession is a woman 
\citep{lucciolistereotipi, vanmassenhove2021gender}.

In this work, we are motivated to address the research question of \textit{how much bias exists in the text data of multiple languages, if at all bias exists in them}?
We particularly investigate 6 benchmark datasets on the English GLUE/SuperGLUE leaderboards \citep{wang-etal-2018-glue,NEURIPS2019_4496bf24} and one dataset each for the other 4 languages: Italian, Dutch, German, and Swedish.
%Their brief details are provided in Section \ref{method}.
First, we train \acrshort{sota} \acrfull{mt5} \citep{xue-etal-2021-mt5} and multilingual Bidirectional Encoder Representations
from Transformers (mBERT) models for bias classification on the \acrfull{mab} for each language, in a similar setup as \citet{ALKHALED2023100030}.
For the evaluations, we search through the literature to compare different metrics or evaluation methods as shown in
Table \ref{table:metcompare} and discussed in Section \ref{litrev}.
This motivates our choice of bipol, the multi-axes bias metric, which we then compare in experiments with a lexica baseline method.
In addition, to confirm the unverified assumption that toxic comments contain bias \citep{sap-etal-2020-social,ALKHALED2023100030}, we annotate 200 randomly-selected samples from the training set of the English \acrshort{mab}.

\begin{table}[h!]
%\tiny
\centering
\resizebox{\columnwidth}{!}{%
%\begin{tabular}{p{0.07\linewidth} p{0.85\linewidth}}
\begin{tabular}{lcc}
\hline
\textbf{Metric/Evaluator} & \textbf{Axis}   & \textbf{Terms}\\
\hline
Winogender \citep{rudinger-etal-2018-gender} & 1 &  60 \\
WinoBias \citep{zhao-etal-2018-gender} & 1  &  40\\
StereoSet \citep{nadeem-etal-2021-stereoset} & 4 & 321\\
%Hurtlex \cite{nozza2021honest} & 6 & 1,072 \\
GenBiT \citep{sengupta2021genbit} & 1 & - \\
CrowS-Pairs \citep{nangia-etal-2020-crows} & 9 & 3,016 \\
Bipol \citep{ALKHALED2023100030} & $>$2,  $<$13 & $>$45,  $<$466\\
 \hline
\end{tabular}
}
\caption{\label{table:metcompare} \footnotesize Comparison of some bias evaluation methods.}
\end{table}

\newpage
\paragraph{Our Contributions}
\begin{itemize}
    \item We make available new large labeled datasets on bias of almost 2 million samples each for 3 languages: Italian, Dutch, and German.\footnote{github.com/LTU-Machine-Learning/bipolmulti}
    
    \item We make available lexica of sensitive terms for bias detection in the 3 languages.
    
    \item We confirm the unverified assumption in the underlying datasets of \acrshort{mab} (\acrfull{sbic} and Jigsaw) \citep{ALKHALED2023100030} that toxic comments contain bias.
\end{itemize}

The rest of this paper is organized as follows.
In Section \ref{litrev}, we discuss the literature review of related work.
In Section \ref{bipol}, we briefly discuss the \textit{bipol} metric.
In Section \ref{method}, we explain the steps involved in the methodology and the datasets we use.
In Section \ref{results}, we present our findings and discuss them.
In Section \ref{conclusion}, we end with the conclusion and possible future work.

\section{Literature Review}
\label{litrev}
%\subsubsection{Bias in Italian}
Although English usually gets more support and attention in the literature, there have been attempts at measuring and mitigating bias in other languages.
Testing for the presence of bias in Italian often has a contrastive perspective with English, with a focus on gender bias \citep{gaido-etal-2021-split, rescigno2020case}.
MuST-SHE \citep{bentivogli-etal-2020-gender} and gENder-IT \citep{vanmassenhove2021gender}
are examples of gender bias evaluation sets.
%Italian exhibits gender marks through inflectional morphology and through syntax by agreement \cite{bentivogli-etal-2020-gender}.
%To test the performance of language models in the tasks of text and speech translation, MuST-SHE, a Gender Bias Evaluation Testset, was developed \cite{bentivogli-etal-2020-gender}.
%Based on MuST-SHE, gENder-IT was also created \cite{vanmassenhove2021gender}.
%Another task that has been tested with respect to gender bias is text generation, for which \citet{nozza2021honest} developed a score (“HONEST”), designed to measure the probability that a language model will output a biased text given a certain template. 
Going beyond gender bias, \citet{kurpicz2021world} and \citet{huang2020multilingual} also identified biases related to people's origin and speakers' age.
It is essential to remember that the mentioned biases can be vehicles for misogynous and hateful discourse \citep{el2020mdd, attanasio2022entropy, merenda2018source}.

%In general, the most tested model is BERT \cite{nozza2021honest, attanasio2020politeam, huang2020multilingual}.
%However, experiments have also been carried out with RNN, CNN \cite{huang2020multilingual} and SVM \cite{attanasio2020politeam, merenda2018source}. \citet{el2020mdd} points out that static embeddings are richer in bias than contextual embeddings. 

%To conclude, work on Italian also sees ongoing research on the regularization of unintended biases, i.e. preventing linguistic models designed for bias identification from reporting an excessive number of false positives. \citet{attanasio2022entropy}, for instance, suggests fine-tuning BERT with an Entropy-based Attention Regularization.

%\subsubsection{Bias in Dutch}
Bias studies for Dutch mostly consider binary gender bias.
\citet{chavez-mulsa-spanakis-2020-evaluating} investigate gender bias in Dutch static and contextualized word embeddings by creating Dutch versions of the Word/Sentence Embedding Association Test (WEAT/SEAT) \citep{caliskan2017semantics,may-etal-2019-measuring}.
WEAT measures bias in word embeddings and can be limited in scope, in addition to having sensitivity to seed words.
\citet{mccurdy2020grammatical} perform a similar evaluation in a multilingual setup to compare the effect of grammatical gender saliency across languages. %Continuing, \citet{delobelle2020robbert} investigate %binary 
%gender occupation bias in robBERT, a Dutch RoBERTa \cite{liu2019roberta} based model, through a template-based association test \cite{kurita-etal-2019-measuring,may-etal-2019-measuring}.
%Other studies investigate biases in downstream tasks %applications through multilingual evaluations: \citet{hovy-etal-2020-sound} consider stylistic biases in commercial machine translation systems and \citet{ghaddar2021context} evaluate name regularity bias in NER systems. Finally,
Several works use different \acrshort{nlp} techniques to evaluate bias in corpora of Dutch news articles \citep{wevers2019using, kroon2020clouded, kroon2021s, fokkens-etal-2018-studying} and literary texts \citep{koolen-van-cranenburgh-2017-stereotypes}.

%\subsubsection{Bias in German}
In \citet{kurpicz2020cultural}, bias is measured with regards to place of origin and gender in German word embeddings using WEAT \citep{caliskan2017semantics}.
%It reports statistically significant bias based on the origin of the name in relation to pleasant and unpleasant words, and statistically significant bias based on gender for the categories of family and career.
%Additionally, statistically significant bias was identified in two other sets: a gender bias in the word categories for different subjects of study, %(e.g. subjects such as special pedagogy, veterinary medicine, ethnology, educational science, and psychology related to female while technical studies such as electrical engineering, mechanical engineering, computer science, microtechnology, and physics related to male) and historical gender bias from the 18th century which is still present in today’s word embeddings. %(e.g. character such as Feeling, Sentiment, Receptiveness, Religiousness, and Understanding related to female while Mind, Rationality, Realisation, Thinking, Knowing, Judging related to male). 
In \citet{kurpicz2021world}, an automatic bias detection method (BiasWords) is presented, through which new biased word sets can be identified by exploring the vector space around the well-known word sets that show bias.
%New word sets affirming the gender bias with respect to subjects of study were identified and statistically significant origin bias related to words of professional success and failure was also identified.
%The German peer-reviews study of \cite{wambsganss2022bias}, found that although their collected corpus of peer-reviews did not reveal significant biases using WEAT, the pre-trained German language models (GermanBERT, German T5, and German GPT-2) had substantial conceptual, racial, and gender bias which increased during fine-tuning on the peer-review data.
%In the work of \cite{kraft2022measuring}, German regard classifier was developed, which measured the gender bias in german language generation models (German GPT-2, GPT-3).
%It revealed positive regard toward female supporting sexist stereotypes (e.g. family- and care-related terms are related to female while crime and perpetrators are related to male).
In the template-based study of \citet{cho2021towards}, on gender bias in translations, the accuracy of gender inference was measured for multiple languages including German.
It was found that, particularly for German, the inference accuracy and disparate impact were lower for female than male, implying that certain translations were wrongly performed for cases that required female inference.
%Also, the difference between occupation and sentiment was found to be small, it suggests that the projection of bias apart from social roles also affects judgments on personality. Work of \cite{folkerts2019analyzing} performs experiments on sentiment analysis (using Google Natural Language API, Amazon Web Service Comprehend, IBM Watson Natural Language Understanding, and Microsoft Azure Cognitive Service) on german job reference letters with varying subjects (e.g. male or female, German or Turkish surnames, and German surnames with or without nobiliary particle) was performed. It was found that the same template can render extremely different sentiments depending on the subject. Work of \cite{zhao2020gender} illustrates ways for quantifying gender bias in multilingual embeddings through intrinsic and extrinsic perspectives.
%The gender distribution in the Multilingual BiosBias Dataset that \citet{zhao2020gender} built, instances were found to be biased towards male with a ratio of 3.53 : 1 for German.
Since German is a grammatically gendered, morphologically rich language, \citet{gonen2019lipstick} found that debiasing methods of \citet{bolukbasi2016man} were ineffective on german word embeddings.
%However, in \cite{zmigrod2019counterfactual} CDA (Counterfactual Data Augmentation) method for bias mitigation measures was presented.
%In addition, bias in machine translation can be mitigated using adversarial learning framework \cite{fleisig2022mitigating}.

%\subsubsection{Bias in Swedish}
For Swedish, the main focus of bias research appears to be on gender.
\citet{SahlgrenO19} show with their experiments that gender bias is present in pretrained Swedish language models. %Contextualized language models seem less sensitive to gender bias compared to word-based embeddings.
%\citet{article2} debiasing method do not work for Swedish, and does, according to  \citet{SahlgrenO19}, worsen the bias in the contextual models (BERT and ELMo).
%\citet{cmsf2022003003} and \citet{edsoaion123525133120190101} 
\citet{cmsf2022003003} and \citet{edsoai.on123525133120190101}
found that the male gender tends to be associated with higher-status professions. %\citet{cmsf2022003003} also found that using larger models may lead to increased gender bias.
A study with data from mainstream news corpora by \citet{edsswe.oai.DiVA.org.uu.43035020200101} shows that women are associated with concepts like family, communication and relationships.
%AI Fairness 360 \cite{bellamy2018ai} and similar libraries are based on first identifying privileged and unprivileged groups (not spans of text like bipol) in order to calculate bias/fairness.
%While men are neutral in the corresponding English corpora, are men in Sweden best associated with the label "crime and punishment". The topic of crime and punishment is present in 40% of the articles  \cite{edsswe.oai.DiVA.org.umu.17758320200101}. 
%Some do not publicly provide their codes for comparison.

\section{Bipol}
\label{bipol}

For the purpose of this work, we summarize \textit{bipol} here but details are discussed in \citet{ALKHALED2023100030}.
The \textit{bipol} metric uses a two-step mechanism for estimating bias in text data: binary classification and sensitive term evaluation using lexica.
It has maximum and minimum values of 1 and 0, respectively.
Bipol is expressed in Equations \ref{eq:eq2} and \ref{eq:eq3} from the main Equation \ref{eq:eq1}, where $ b_{c} $ is the classification component and $ b_{s} $ is the sensitive term evaluation component.

\begin{subequations}
\small

\begin{equation}
\mathit{b}=\begin{cases}
\mathit{b_{c}} . \mathit{b_{s}}, & \text{if $b_{s}>0$}\\
\mathit{b_{c}}, & \text{otherwise}
\end{cases}
\label{eq:eq1}
\end{equation}

%\begin{equation}
%\mathit{b} = 
%\mathit{b_{c}} . %\mathit{b_{s}}
%\label{eq:eq1}
%\end{equation}

\begin{equation}
\mathit{b_{c}} = 
\frac{tp + fp}{tp+fp+tn+fn}
\label{eq:eq2}
\end{equation}

\begin{equation}
\mathit{b_{s}} = \frac{1}{r} \sum_{t=1}^{r} 
{\left( \frac{1}{q} \sum_{x=1}^{q} {\left(
\frac{|\sum_{s=1}^{n} a_{s} - \sum_{s=1}^{m} c_{s}|}{\sum_{s=1}^{p} d_{s}}
\right)}_{x}
\right)}_{t}
\label{eq:eq3}
\end{equation}
\end{subequations}

In step 1, a trained model is used to classify all the samples.
The ratio of the biased samples to the total samples predicted is determined.
The \textit{\textbf{tp}}, \textit{\textbf{fp}}, \textit{\textbf{tn}}, and \textit{\textbf{fn}} are values of the true positives, false positives, true negatives, and false negatives, respectively.
Since there's hardly a perfect classifier, the positive error rate is usually reported.
False positives are known to exist in similar classification systems like spam detection and automatic hate speech detection \citep{heron2009technologies,feng2018multistage}.
%sci5040037
    
Step 2 is similar to \textit{term frequency-inverse document frequency} (TF-IDF) in that it is based on term frequency \citep{salton1988term,ramos2003using}, 
Biased samples from step 1 are evaluated token-wise along all possible bias axes, using all the lexica of sensitive terms.
An axis is a domain such as gender or race.
Tables \ref{table:lexi} and \ref{table:lexi2} provide the lexica sizes.
For English and Swedish, we use the same lexica released by \citet{ALKHALED2023100030} and \citet{adewumi2023bipol}, respectively.
For the other 3 languages, we create new lexica of terms (e.g. she \& her) associated with specific gender or stereotypes from public sources.\footnote{fluentu.com/blog/italian/italian-nouns, en.wiktionary.org/wiki/Category:Italian\_offensive\_terms, Dutch\_profanity, Category:German\_ethnic\_slurs}
Some of the terms in the lexica were selected from the sources based on the topmost available.
These may also be expanded as needed, since bias terms are known to evolve \citep{haemmerlie1991goldberg,antoniak2021bad}.
The non-English lexica are small because fewer terms are usually available in other languages compared to the high-resource English language and we use the same size across the languages to be able to compare performance somewhat.
The Appendix lists these terms.

Equation \ref{eq:eq3} first finds the absolute difference between the two maximum summed frequencies in the types of an axis (\(|\sum_{s=1}^{n} a_{s} - \sum_{s=1}^{m} c_{s}| \)), where \textit{n} and \textit{m} are the total terms in a sentence along an axis. For example, in the sentence ´\textit{Women!!! PERSON taught you better than that. Shame on you!}', female terms = 1 while male terms = 0.
This is then divided by the summed frequencies of all the terms ($d_{s}$) in that axis (\( \sum_{s=1}^{p} d_{s} \)).
The operation is performed for all axes ($q$) and the average taken (\( \frac{1}{q} \sum_{x=1}^{q} \)).
It is performed for all the biased samples ($r$) and the average taken (\( \frac{1}{r} \sum_{t=1}^{r} \) ).

% Women are that way, so it's not shocking she slapped him

\begin{table}[h!]
%\small
\centering
\resizebox{\columnwidth}{!}{%
%\begin{tabular}{p{0.07\linewidth} p{0.85\linewidth}}
\begin{tabular}{lccc}
\hline
\textbf{Axis} & \textbf{Type 1} & \textbf{Type 2} & \textbf{Type 3}\\
\hline
Racial & 84 (black) & 127 (white) & \\
Gender & 76 (female) & 46 (male) & \\
Religious & 180 (christian) & 465 (muslim) & 179 (hindu)\\
\hline
\end{tabular}
}
\caption{\label{table:lexi}
English lexica sizes \citep{ALKHALED2023100030}. \footnotesize An example of racial-black is \textit{negro}.
}
\end{table}

\begin{table}[h!]
\small
\centering
%\resizebox{\columnwidth}{!}{%
%\begin{tabular}{p{0.07\linewidth} p{0.85\linewidth}}
\begin{tabular}{lcc}
\hline
\textbf{Axis} & \textbf{Type 1} & \textbf{Type 2} \\
\hline
Gender & 17 (female) & 17 (male)\\
Racial & 10 (black) & 10 (white)\\
%Religious & 180 (christian) & 465 (muslim) & 179 (hindu)\\
\hline
\end{tabular}
%}
\caption{\label{table:lexi2}
Italian, Dutch, German and Swedish Lexica sizes. \footnotesize An example of gender-female is \textit{woman}.
}
\end{table}

\section{Methodology}
\label{method}

\subsection{Datasets}
\subsubsection{\acrshort{mab}}

\begin{table*}[h]
%\small
\centering
%\resizebox{\columnwidth}{!}{%
%\begin{tabular}{p{0.07\linewidth} p{0.85\linewidth}}
\begin{tabular}{p{0.55\linewidth} | p{0.1\linewidth} | 
 p{0.09\linewidth} | p{0.1\linewidth}}
\hline
\textbf{comment\_text} & \textbf{label} & \textbf{old\_id} & \textbf{id}  %\textbf{axis}
\\
\hline
A fitness ranking that drops each month as more and more fat people move to Colorado. & biased & 6231681 & 167
\\
\hline
Een fitness ranking die elke maand daalt als meer en meer dikke mensen verhuizen naar Colorado. & & & 
\\
\hline
Ein Fitness-Ranking, das jeden Monat fällt, wie mehr und mehr fette Menschen nach Colorado zu bewegen. & & & 
\\
\hline
Un ranking di fitness che scende ogni mese come sempre più persone grasse si spostano in Colorado.
 & & & 
\\
\hline
En fitness ranking som sjunker varje månad som fler och fler feta människor flyttar till Colorado.
 & & & 
\\
\hline
& & & 
\\
\hline
\hline
Women ! ! ! PERSON taught you better than that. Shame on you! & biased & 6233025 & 628
\\
\hline
Vrouwen ! ! ! PERSOON heeft je beter geleerd dan dat. Je moet je schamen!
& & &
\\
\hline
Frauen!!!! PERSON lehrte Sie besser als das. Schande über Sie!
 & & & 
\\
\hline
Donne ! ! ! Person ti ha insegnato meglio di così, vergognati!
 & & & 
\\
\hline
Kvinnor ! ! !- Han lärde dig bättre än så.  Skäms på dig!
 & & & 
\\
\hline
 & & & 
\\
\hline
\hline
\end{tabular}
%}
\caption{\label{table:newbiasdata}
\textbf{English}, \textbf{Dutch}, \textbf{German}, \textbf{Italian}, and \textbf{Swedish} examples from the \acrshort{mab} dataset. "PERSON" is the anonymization of a piece of \acrfull{pii} in the dataset.
}
\end{table*}

\begin{table}[h]
%\small
\centering
%\resizebox{\columnwidth}{!}{%
%\begin{tabular}{p{0.07\linewidth} p{0.85\linewidth}}
\begin{tabular}{lccc}
\hline
\textbf{Set} & \textbf{Biased} & \textbf{Unbiased} & \textbf{Total}\\
\hline
Training & 533,544 & 1,209,433 & 1,742,977 \\
Validation & 32,338 & 69,649 & 101,987\\
Test & 33,470 & 68,541 & 102,011\\
\hline
 & 599,352 & 1,347,623 & 1,946,975 \\
 \hline
\end{tabular}
%}
\caption{\label{table:genbias}
\acrshort{mab} dataset split
}
\end{table}

The Italian, Dutch and German datasets were machine-translated from \acrshort{mab}\footnote{The reference provides details of the annotation of the base data.} %\citep{ALKHALED2023100030}
with the high-quality Helsinki-\acrshort{nlp} model \citep{TiedemannThottingal:EAMT2020}.
Each translation took about 48 hours on one GPU.
Examples from the data are provided in Table \ref{table:newbiasdata}.
%The features in the two datasets are, hence, the same.
Table \ref{table:genbias} provides statistics about the datasets.
For quality control (QC), we verified translation by back-translating some random samples using Google \acrshort{nmt}.
Personal identifiable information (\acrshort{pii}) was removed from the \acrshort{mab} dataset using the spaCy library.
The 3 datasets are used to train new bias classifiers.
We also train on the original English and the Swedish. %\cite{adewumi2023bipol}.

\paragraph{Machine-Translation issues:}
Culture-specific biases may not be represented in the \acrshort{mab} versions for the translated languages because the original dataset is in English.
This is a limitation.
However, bias is also a universal concern, such that there are examples that span across cultures.  
For instance, the examples in Table \ref{table:newbiasdata} are of universal concern because individuals with non-conforming bodies and women should be respected, regardless of culture or nationality.
Hence, the \acrshort{mab} versions are relevant for bias detection, though they were translated.

\subsubsection{Evaluation datasets}
Ten datasets are evaluated for bias in this work.
All are automatically preprocessed before evaluation, the same way the training data were preprocessed.
This includes removal of IP addresses, emojis, URLs, special characters, emails, extra spaces, numbers, empty text rows, and duplicate rows.
All texts are then lowercased.

We selected datasets that are available on the HuggingFace \citep{wolf-etal-2020-transformers} Datasets.
We evaluated the first 1,000 samples of each training split due to resource constraints.
The understanding is that if bias is detected in these samples, then scaling over the entire dataset means there's probability of more bias. 
For English, we evaluated the sentence column of \acrfull{cola} \citep{warstadt2019neural}, the sentence column of \acrfull{qnli} \citep{wang-etal-2018-glue}, the sentennce1 column of \acrfull{mrpc} \citep{dolan-brockett-2005-automatically}, the premise column of \acrfull{mnli} \citep{williams-etal-2018-broad}, the premise column of the \acrfull{cb} dataset \citep{de2019commitmentbank}, and the passage column of \acrfull{record} \citep{zhang2018record}.
For Italian, we evaluated the context column of the \acrfull{squad} \citep{10.1007/978-3-030-03840-3_29,rajpurkar-etal-2016-squad};
for Dutch, the sentence1 column of the \acrfull{stsb} \citep{cer-etal-2017-semeval}; for  German, the text column of the \acrfull{gnad10} \citep{Schabus2017};
for Swedish, the premise of the \acrshort{cb}.
%More details of the datasets will be available in the appendix.

\subsection{Annotation for the assumption confirmation}
To verify the assumption that toxic comments contain bias,
we randomly selected 200 samples from the training set of \acrshort{mab}-English for annotation on Slack, an online platform.
The selection of 200 samples is based on an error margin of 7\% and a confidence level of 95\%.
To ensure high-quality annotation, we use established techniques for this task: 1) the use of gold (30) samples, 2) multiple (i.e. 3) annotators, and 3) minimum qualification of undergraduate study for annotators.
Each annotator was paid 25 U.S. dollars and the it took about 2 hours to complete the annotation on average.
We mixed the 30 gold samples with the 200, to verify the annotation quality of each annotator, as they were required to get, at least, 16 correctly for their annotation to be accepted.
The 30 gold samples are samples with unanimous agreement in the original Jigsaw or \acrshort{sbic} data, which make up the \acrshort{mab}.
We provide \acrfull{iaa} using Jaccard similarity coefficient (intersection over union) and \acrfull{cus} \citep{10191208} (intersection over sample size).
%The annotation instruction is provided in the appendix.
%We collected the work of 3 valid annotators to establish the \acrfull{iaa}.

\subsection{Experiments}

We selected two \acrfull{sota} pre-trained, multilingual models for experiments to compare their macro F1 performance: \acrshort{mt5}-small and m\acrshort{bert}-base.
These are from the HuggingFace hub.
We further report the \acrshort{mt5} positive error rate of predictions.
The \acrshort{mt5}-small has 300 million parameters \citep{xue-etal-2021-mt5} while m\acrshort{bert}-Base has 110 million parameters.
We trained only on the \acrshort{mab} datasets and evaluated using only the \acrshort{mt5} model, the better model of the 2, as will be observed in Section \ref{results}.
For the \acrshort{cb} and \acrshort{record} datasets, we evaluate all samples since they contain only about 250 and 620 entries, respectively.
We used wandb \citep{wandb} for hyper-parameter exploration, based on Bayesian optimization.
For \acrshort{mt5}, we set the maximum and minimum learning rates as 5e-5 and 2e-5 while the maximum and minimum epochs are 20 and 4, respectively.
One epoch is equivalent to the ratio of the total number of samples to the batch size (i.e. the steps).
We used a batch size of 8 because higher numbers easily resulted in memory challenges.

For m\acrshort{bert}, we set the learning rates and epochs as with \acrshort{mt5}.
However, we explore over batch sizes of 8, 16 and 32.
For both models, we set the maximum input sequence length to 512.
Training took, on average, about 7.3 hours per language per epoch for m\acrshort{bert} while it was 6 hours for \acrshort{mt5}.
For all the experiments, we limit the run counts to 2 per language because of the long training time each takes on average.
The average scores of the results are reported.
The saved models with the lowest losses were used to evaluate the datasets.
All the experiments were performed on two shared Nvidia DGX-1 machines that run Ubuntu 20.04 and 18.04.
One machine has 8 x 40GB A100 GPUs while the other has 8 x 32GB V100 GPUs.

The lexica baseline, compared in experiments, is similar to the equation of the second step in bipol.
It does not consider bias semantically and uses term frequencies, similarly to TF-IDF.
It uses the same lexica as bipol.
Its maximum and minimum values are 1 and 0, respectively.

\section{Results and Discussion}
\label{results}

From Table \ref{table:res1}, we observe that all \acrshort{mt5} results are better than those of m\acrshort{bert} across the languages.
The two-sample t-test of the difference of means between all the corresponding \acrshort{mt5} and m\acrshort{bert} scores have \textit{p} values $<$ 0.0001 for alpha of 0.05, showing the results are statistically significant.
It appears better hyper-parameter search may be required for the m\acrshort{bert} model to converge and achieve better performance.
The best macro F1 result is for English \acrshort{mt5} at 0.787.
This is not surprising, as English has the largest amount of training data for the pre-trained \acrshort{mt5} model \citep{xue-etal-2021-mt5}.
This occurred at the learning rate of 2.9e-5 and step 1,068,041.
% Significance test

\begin{table}[h]
%\small
\centering
\resizebox{\columnwidth}{!}{%
%\begin{tabular}{p{0.07\linewidth} p{0.85\linewidth}}
\begin{tabular}{cccc}
\hline
 & \multicolumn{2}{c}{\textbf{macro F1} $\uparrow$ (s.d.)} & \textbf{\acrshort{mt5}} \textbf{error} $\downarrow$\\
%\hline
\textbf{\acrshort{mab} version} &
\textbf{m\acrshort{bert}} &
\textbf{\acrshort{mt5}} & \textbf{\acrshort{fp}/(\acrshort{fp}+\acrshort{tp}})
\\
\hline
English & 0.418 (0.01) & 0.787 (0) & 0.261 \\
Italian & 0.429 (0) & 0.768 (0) & 0.283 \\
Dutch & 0.419 (0.01) & 0.768 (0) & 0.269\\
German & 0.418 (0.01) & 0.769 (0) & 0.261 \\
Swedish & 0.418 (0.01) & 0.768 (0) & 0.274
 \\ \hline
\end{tabular}
}
\caption{\label{table:res1}
Average F1 scores on the validation sets.
}
\end{table}

\begin{table}[h!]
%\small
\centering
\resizebox{\columnwidth}{!}{%
\begin{tabular}{ccccc}
\hline
 & \multicolumn{3}{c}{\textbf{bipol scores}} $\downarrow$ (s.d.) & \\
%\hline
\textbf{English} & \textbf{$\mathit{b_{c}}$} & \textbf{$\mathit{b_{s}}$} & \textbf{bipol \textit{(b)}} & \textbf{baseline} $\downarrow$
\\
\hline
\acrshort{cb} & 0.096 & 0.875 & 0.084 (0) & 0.88\\
\acrshort{cola} & 0.101 & 0.943 & 0.095 (0) & 0.958\\
\acrshort{record} &  0.094 & 0.852 & 0.025 (0) & 0.829\\
\acrshort{mrpc} &  0.048 & 0.944 & 0.045 (0) & 0.957\\
\acrshort{mnli} &  0.063 & 0.833 & 0.053 (0) & 0.965\\
\acrshort{qnli} &  0.03 & 0.933 & 0.028 (0) & 0.945\\
\hline
Italian  & & & & \\
%\hline
%CoLA &  &  & (0) & 0.999\\
SQuAD & 0.014 & 0 & 0.014 (0) & 0.989\\
\hline
Dutch  & & & & \\
%\hline
\acrshort{stsb} & 0.435 & 0.992 & 0.432 (0) & 0.987\\
%DailyMail &  &  &  (0) & 0.999\\\\
 \hline
German  & & & & \\
%\hline
%Open Legal &  &  & (0) & 0.993\\
\acrshort{gnad10} & 0.049 & 0.502 & 0.025 (0) & 1\\
\hline
Swedish  & &  & & \\
%\hline
\acrshort{cb} & 0.08 & 0.938 & 0.075 (0) & 0.97\\
%DIACRON  &  &  & (0) & 0.994\\
\hline
\end{tabular}
}
\caption{Average bipol \& lexica baseline scores.}
\label{table:res2}
\end{table}

\begin{figure*}[h]
\begin{center}
\includegraphics[scale=0.08]{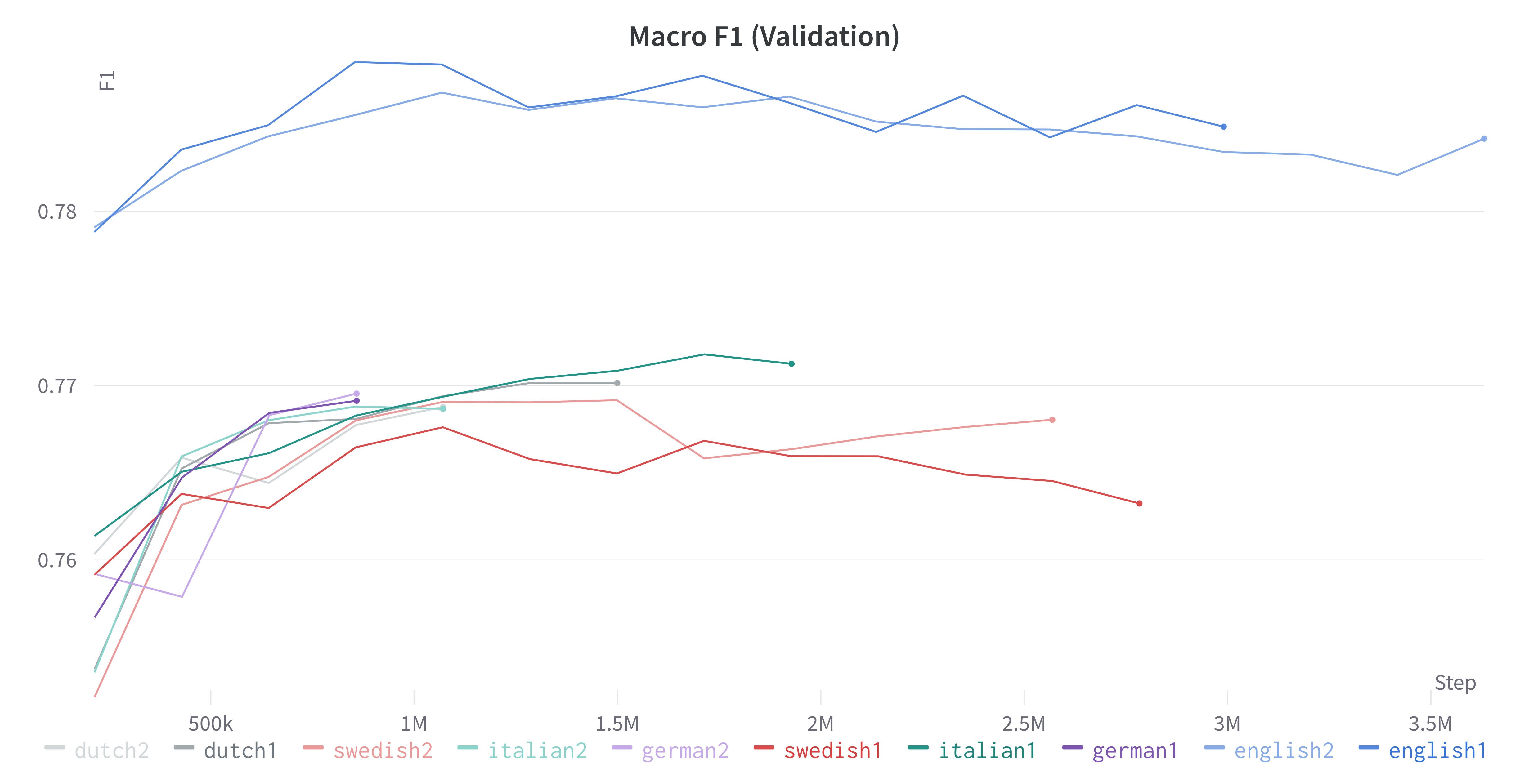} 
\caption{Macro F1 of the validation set for the 5 languages, as generated by wandb.}
\label{fig1}
\end{center}
\end{figure*}

\begin{figure*}[h]
\begin{center}
\includegraphics[scale=0.08]{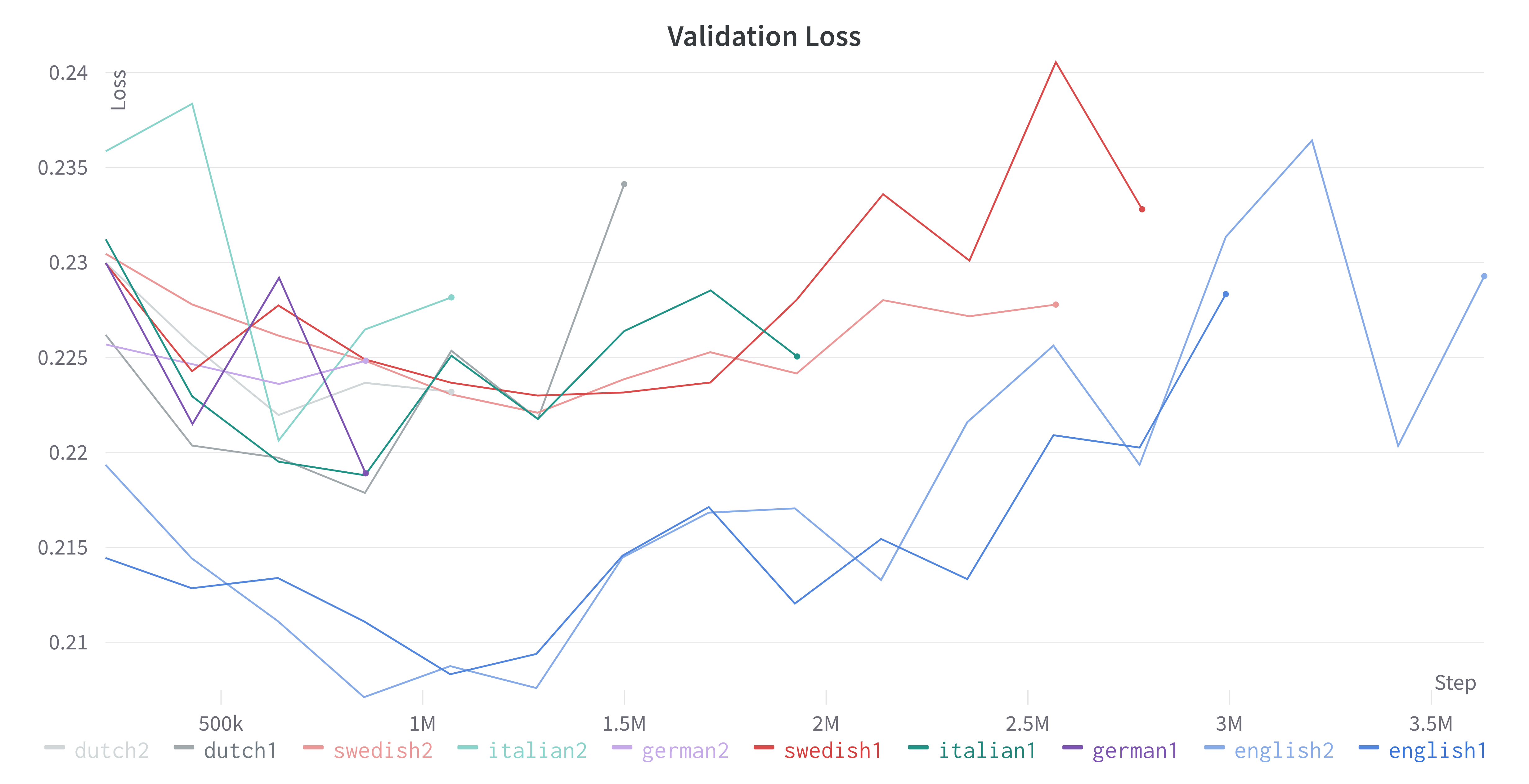} 
\caption{Loss on the validation set for the 5 languages, as generated by wandb.}
\label{fig2}
\end{center}
\end{figure*}

Figures \ref{fig1} and \ref{fig2} depict the validation sets macro F1 and loss line graphs for the 2 runs for the 5 languages, respectively.
From Table \ref{table:res2}, we observe that all the evaluated datasets have biases, though seemingly little (but important) when compared to the maximum of 1.
We say important because many of the datasets contain small number of samples yet they can be detected.
Furthermore, a low value does not necessarily diminish the weight of the effect of bias in society or the data but we leave the discussion about what amount should be tolerated open for the \acrshort{nlp} community.
Our recommendation is to have a bias score as close to zero as possible.
On the other hand, the lexica baseline appears overly confident of much more bias, which is incorrect because the method fails to exclude unbiased text in its evaluation, which is a shortcoming of methods based solely on it.
The Dutch STSB is higher than the other bipol scores because of the higher bipol classifier component score of 0.435, which may be because of the nature of the dataset.

\begin{figure}[h!]
\centering
\includegraphics[width=0.4\textwidth]{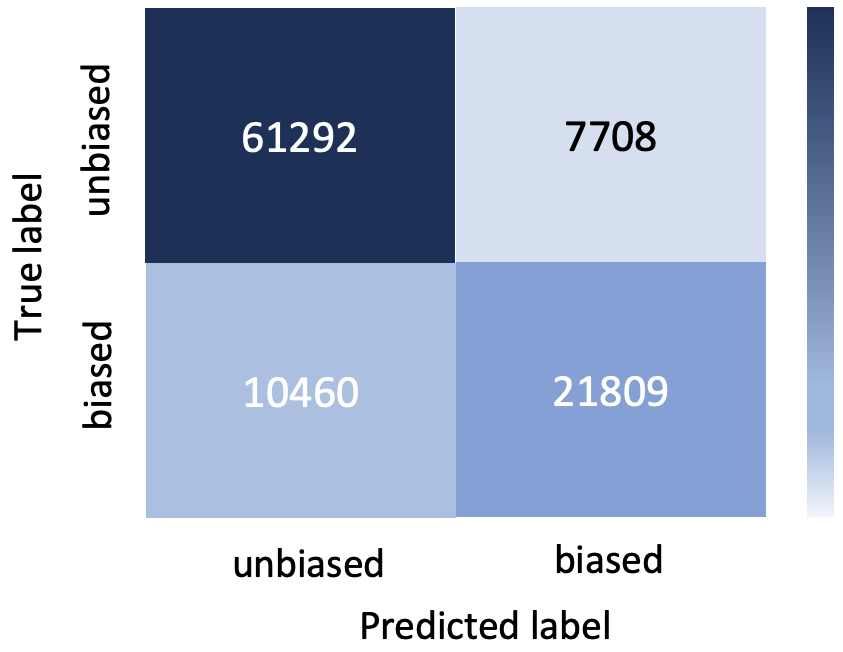}
\caption{Error matrix of \acrshort{mt5} on \acrshort{mab}-English}
\label{error_mat}
\end{figure}

\subsection{Error analysis \& qualitative results}

According to the error matrix in Figure \ref{error_mat}, the \acrshort{mt5} model is better at correctly predicting unbiased samples.
This is because of the higher unbiased samples in the training data of \acrshort{mab}.
In Table \ref{table:qexamples}, the first example for the English \acrshort{cb} contains a stereotypical statement "\textit{men are naturally
right and it is the role of women to follow their lead}", leading to the correct biased prediction by the model.
Similarly, this correct prediction is made in the Swedish \acrshort{cb}.
%\footnote{It is not shown here due to page limit.}
We notice over-generalization \citep{may-etal-2019-measuring,nadeem-etal-2021-stereoset} in the correct examples for the \acrshort{cola} predictions, where "\textit{every}" is used.
The table also shows some incorrect predictions.
%The complete file artefacts are available for inspection.

\begin{table*}[h!]
%\small
\centering
%\resizebox{\columnwidth}{!}{%
%\begin{tabular}{p{0.07\linewidth} p{0.85\linewidth}}
\begin{tabular}{p{0.58\linewidth} | p{0.12\linewidth} | p{0.1\linewidth}}
\hline
\textbf{comment\_text} & \textbf{prediction} & \textbf{csv id}  %\textbf{axis}
\\
\hline
\multicolumn{3}{c}{\textbf{Correct predictions from the English \acrshort{cb}}}
\\
\hline
nicky approached her with the assumption that men are naturally right and it is the role of women to follow their lead. constance whose confidence was growing daily was not prepared to give in to nickys wishes merely because of his sex. if she felt he was right then she agreed with him. & biased & 85 \\
\hline
b: thats true. a: so. b: uh the other argument is that the death penalty is a deterrent and i really dont uh agree with that. i dont think anyone who would commit uh a crime that would get them the death penalty would stop at the moment and say well i was about to kill and dismember this person but oh if they catch me theyre going to kill me so i better not do it. i just dont think uh that it works that way. a: yeah. i dont think its done.
 & biased & 133\\
\hline
\multicolumn{3}{c}{\textbf{Incorrect prediction}}
\\
\hline
b: yeah and the ground will filter some of it but not all of it. a: no not when you figure i didnt realize one cow produces that much manure
 & biased & 137
\\
\hline
& & 
\\
\hline
\multicolumn{3}{c}{\textbf{Correct predictions from the English \acrshort{cola}}}
\\
\hline
if you give him enough opportunity every senator will succumb to corruption.
 & biased & 266 \\
\hline
every senator becomes more corrupt the more lobbyists he talks to.
 & biased & 277\\
\hline
\multicolumn{3}{c}{\textbf{Incorrect prediction}}
\\
\hline
bill squeezed the puppet through the hole.
 & biased & 82
\\
\hline
& & 
\\
\hline
\end{tabular}
%}
\caption{\label{table:qexamples}
Qualitative examples of apparently correct and incorrect predictions in some of the English datasets, based on the \acrshort{mt5} model.
}
\end{table*}

\subsection{Consistent prediction with perturbation}
An interesting property of relative consistency that we observed with the model predictions, as demonstrated with the \acrshort{cola} dataset, is that when sentences are perturbed, the model mostly maintains its predictions, as long as the grounds for prediction (in this case - over-generalization) remain the same.
The perturbations are inherent in the CoLA dataset itself, as the dataset is designed that way.
Some examples are provided in Table \ref{table:consistency} in the Appendix, where 6 out of 8 are correctly predicted.
This property is repeated consistently in other examples not shown here.

\subsection{Explainability by graphs}
\begin{figure}[h!]
\centering
\includegraphics[width=0.45\textwidth]{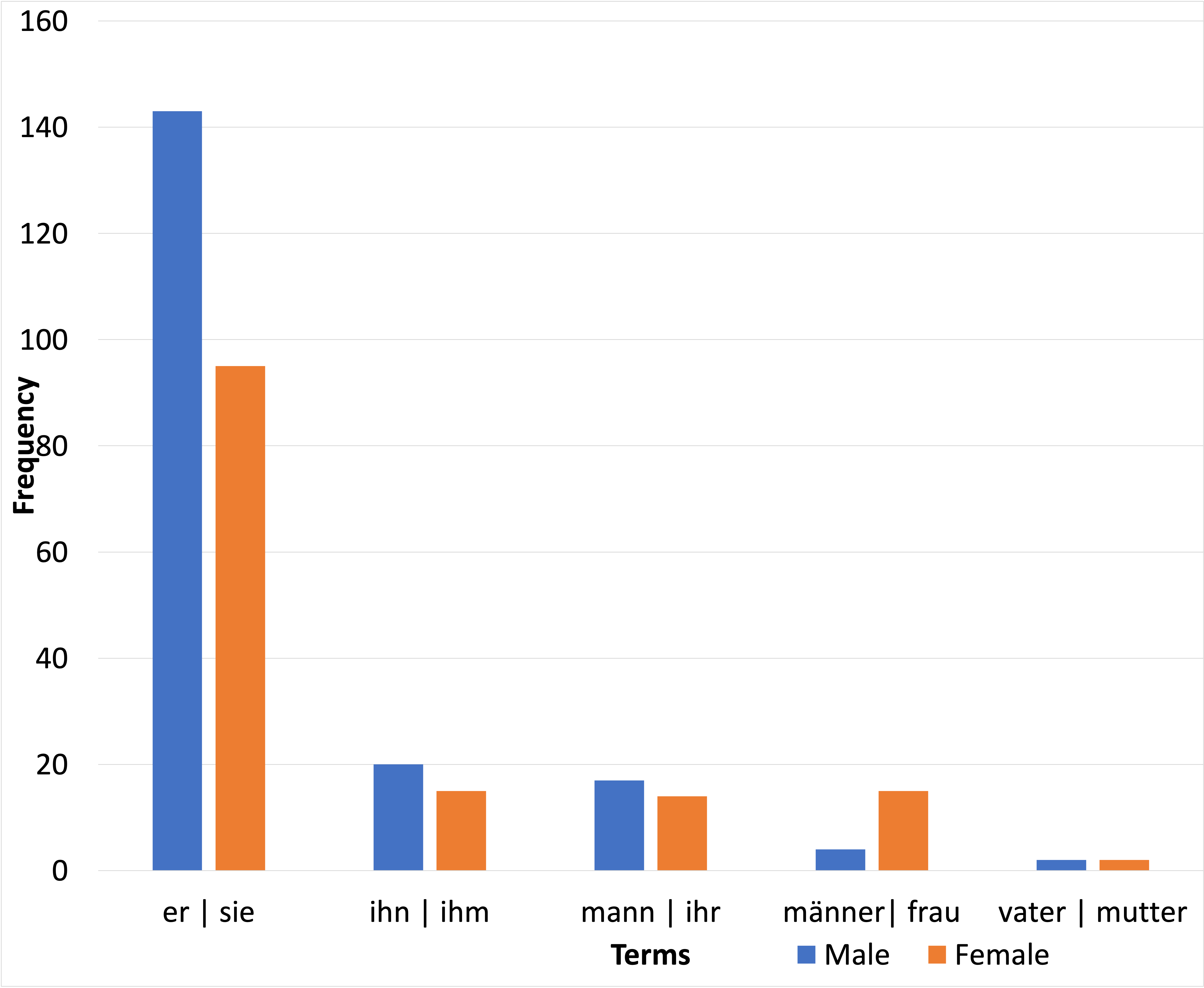}
\caption{Top 5 frequent terms in the \acrshort{gnad10} dataset (\footnotesize paired terms are only for comparison).}
\label{top5gm}
\end{figure}

We show explainability by visualization using graphs.
Bipol produces a dictionary of lists for every evaluation and we show the \textit{top-5 frequent terms} bar graph for the \acrshort{gnad10} dataset in Figure \ref{top5gm}, which has overall male bias.
Many of the 10 evaluated datasets display overall male bias.

%\begin{figure}[h!]
%\centering
%\includegraphics[width=0.5\textwidth]{eng_cola.png}
%\caption{Top 5 frequent terms in the \acrshort{cola} dataset (\footnotesize paired terms are only for comparison).}
%\label{top5en}
%\end{figure}

\subsection{Assumption confirmation through annotation}
\begin{figure}[h!]
\centering
\includegraphics[width=0.5\textwidth]{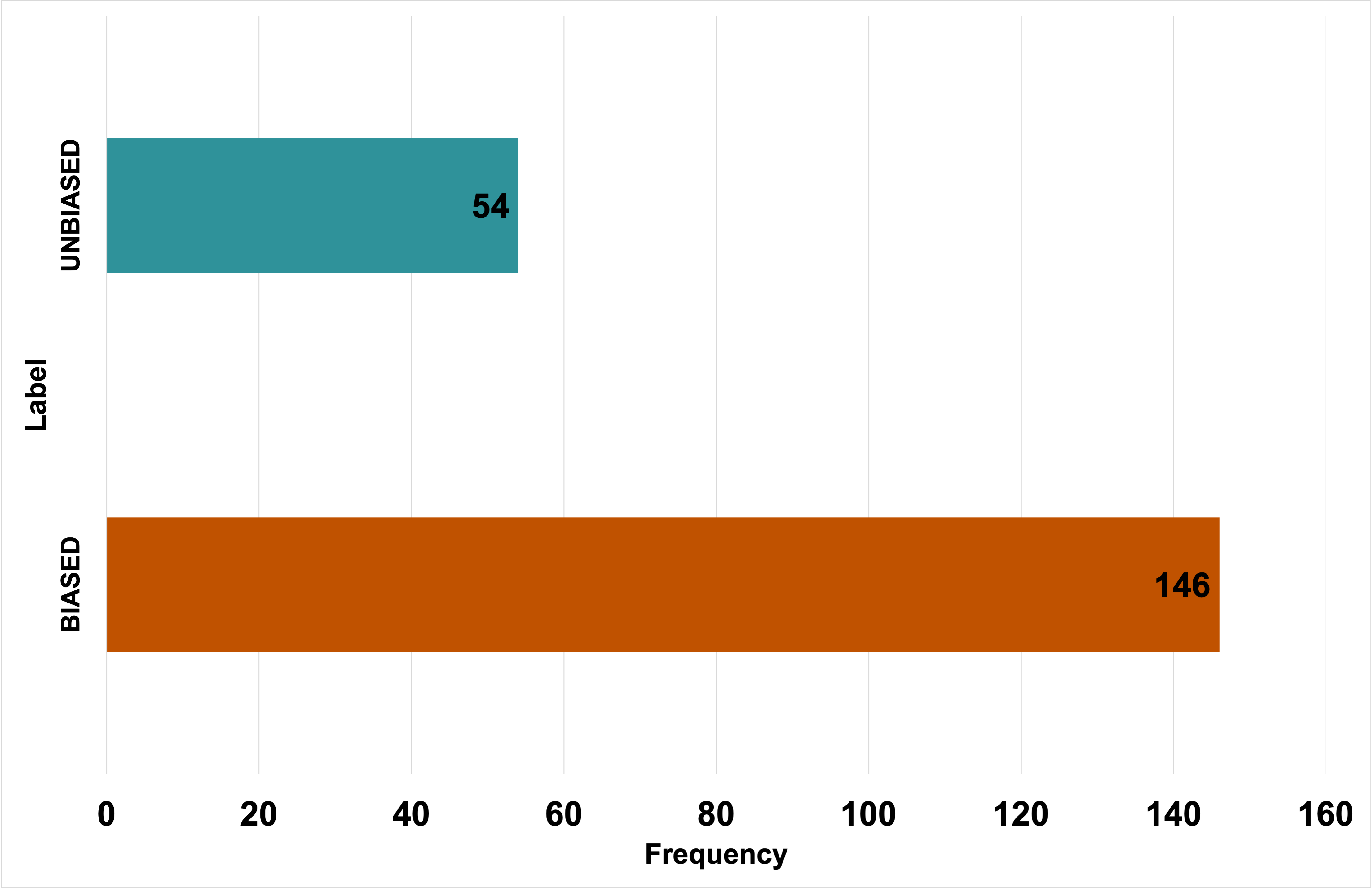}
\caption{Annotation confirms assumption about toxic comments.}
\label{annot}
\end{figure}

The results of the annotation of the 200 \acrshort{mab} samples reveal that toxic comments do contain bias.
This is shown in Figure \ref{annot}.
The Jaccard similarity coefficient and \acrshort{cus} of \acrshort{iaa} are 0.261\footnote{Not to be interpreted using Kappa for 2 annotators on 2 classes. Ours involved 3 annotators} and 0.515, respectively, given that over 50\% is the intersection of unanimous decision.

\section{Conclusion}
\label{conclusion}

The findings of this work show that bias besets \acrfull{nlp} datasets regardless of language, including benchmark datasets on the GLUE/SuperGLUE leaderboards.
We introduced \acrshort{mab} datasets in 3 languages for training models in bias detection.
Each has about 2 million labeled samples.
We also contribute lexica of bias terms for the languages.
In addition, we verified the assumption that toxic comments contain bias.
It may be impossible to completely remove bias from data or models, since they reflect the real world, but resources for estimating bias can provide insight into mitigation strategies for reducing bias.
Future work may explore ways of minimizing false positives in classifiers to make them more effective.
One may also explore how this work scales to other languages or how multilignual models compare to language-specific monolingual models or large language models (\acrshort{llm}s).
Regarding culture-specific biases in datasets, one solution will be to collect data from the specific cultures/nationalities to capture these biases.

\section*{Ethics statement \& limitation}

The authors took care while providing examples of data samples, despite some containing stereotypes or toxic content.
The classifiers trained for estimating the biases in this work are limited in effectiveness, as shown in the results, hence a result of 0 on any dataset does not necessarily indicate a bias-free dataset.
The original \acrshort{mab} was annotated by humans who may have their personal biases based on cultural or demographic  backgrounds.
This means the final annotations may not be seen as absolute ground truth of social biases.
%Finally, effort was made to mask examples with offensive content in this paper.

\subsubsection*{Acknowledgments}
The authors wish to thank the anonymous reviewers for their valuable feedback.
This work is partially supported by the Wallenberg AI, Autonomous Systems and Software Program (WASP), funded by the Knut and Alice Wallenberg Foundation and counterpart funding from Luleå University of Technology (LTU).
The authors also thank Björn Backe for his insightful comments during the writing of this paper. 
%\nocite{Ando2005,augenstein-etal-2016-stance,andrew2007scalable,rasooli-tetrault-2015,goodman-etal-2016-noise,harper-2014-learning}

\bibliography{anthology,custom}
\bibliographystyle{acl_natbib}

\clearpage
\newpage
\appendix

\section{Appendix}
\label{sec:appendix}

\subsection*{Swedish lexica and the English equivalent}

\begin{enumerate}
    \item Gender-female\\
  \begin{inparaenum}[i)]
    \item hon (she)
    \item hennes (her)
    \item flicka (girl)
    \item mor (mother)
    \item kvinna (woman)
    \item dotter (daughter)
    \item mormor (grandmother)
    \item dam (lady)
    \item sondotter (son's daughter)
    \item dotterdotter (daughter's daughter)
    \item tjej (girl)
    \item tjejer (girls)
    \item gumma (old woman)
    \item fru (wife)
    \item slampa (slut)
    \item slyna (slut)
    \item lebb (lesbian)

  \end{inparaenum}
    \item Gender-male \\
      \begin{inparaenum}[i)]
  \item han (he)
    \item pojke (boy)
    \item kille (boy)
    \item far (father)
    \item farfar (grandfather)
    \item hans (his)
    \item man (man)
    \item son (son)
    \item make (husband)
    \item herre (lord)
    \item herrar (lords)
    \item sonson (son's son)
    \item dotterson (daughter's son)
    \item gubbe (old man)
    \item farbro (uncle)
    \item broson (nephew)
    \item män (men)
  \end{inparaenum}
  
    \item Racial-black \\
      \begin{inparaenum}[i)]
  \item neger (nigger)
    \item blåneger (nigger)
    \item blåland (nigger)
    \item blatte (dark immigrants)
    \item svartskalle (black head)
    \item turk (non-Swedish)
    \item sosse (political slur)
    \item svarting (blackness)
    \item partysvenska (political slur)
    \item nigga (nigger)
  \end{inparaenum}
  
    \item Racial-white \\
      \begin{inparaenum}[i)]
  \item svenne (Swedish stereotype)
    \item turk (non-Swedish)
    \item jugge (derogatory term)
    \item sosse (political slur)
    \item hurrare (Finnish Swedish)
    \item lapp (Sami people)
    \item Ang mo (red hair)
    \item partysvenska (political slur)
    \item Ann (White woman)
    \item rutabaga (Swede)
  \end{inparaenum}

\end{enumerate}

\vspace{240pt}
\subsection*{Example of predictions for the CoLA dataset}
\begin{table}[h!]
%\small
\centering
%\resizebox{\columnwidth}{!}{%
%\begin{tabular}{p{0.07\linewidth} p{0.85\linewidth}}
\begin{tabular}{p{0.52\linewidth} | p{0.18\linewidth} | p{0.12\linewidth}}
\hline
\textbf{comment\_text} & \textbf{prediction} & \textbf{csv id}  %\textbf{axis}
\\
\hline
if you give him enough opportunity every senator will succumb to corruption. & biased & 266
\\ \hline
you give him enough opportunity and every senator will succumb to corruption. & biased & 267
\\
\hline
we gave him enough opportunity and sure enough every senator succumbed to corruption. & unbiased & 268
 \\ \hline
if you give any senator enough opportunity he will succumb to corruption. & biased & 269
 \\
\hline
you give any senator enough opportunity and he will succumb to corruption. & biased & 270
\\
\hline
you give every senator enough opportunity and he will succumb to corruption.& biased & 271
 \\ \hline
we gave any senator enough opportunity and sure enough he succumbed to corruption.& biased & 272
\\
\hline
we gave every senator enough opportunity and sure enough he succumbed to corruption. & unbiased & 273
\\ \hline
\end{tabular}
%}
\caption{\label{table:consistency}
Mostly consistent correct prediction with perturbation in the \acrshort{cola} dataset.
}
\end{table}

\end{document}